\documentclass{article}

\usepackage{fullpage}
\usepackage{natbib}
\usepackage{microtype}
\usepackage{graphicx}
\usepackage{subfigure}
\usepackage{booktabs} % for professional tables
\usepackage{graphicx}
\usepackage{subfigure}
\usepackage[most]{tcolorbox}
\usepackage{caption}
\usepackage{xfrac}

\usepackage{hyperref}

\usepackage{algorithmic}
\usepackage{amsmath}
\usepackage{amssymb}
\usepackage{mathtools}
\usepackage{amsthm}

\usepackage[capitalize,noabbrev]{cleveref}

\theoremstyle{plain}
\newtheorem{theorem}{Theorem}[section]

\newtheorem{corollary}[theorem]{Corollary}
\theoremstyle{definition}
\newtheorem{definition}[theorem]{Definition}
\newtheorem{assumption}[theorem]{Assumption}
\theoremstyle{remark}

\usepackage[textsize=tiny]{todonotes}
\usepackage{bm}

\newcommand{\state}{\theta}
\newcommand{\statespace}{\Theta}
\newcommand{\vstate}{\bm{\state}}

\newcommand{\score}{S}

\newcommand{\report}{r}
\newcommand{\vreport}{\bm{\report}}
\newcommand{\rspace}{R}

\newcommand{\mean}{\mu}
\newcommand{\vmean}{\bm{\mean}}

\newcommand{\expect}[2]{{\mathbf{E}}_{#1}\left[#2\right]}

\newcommand{\npointrubric}{n}
\newcommand{\ndim}{m}

\DeclareMathOperator*{\argmax}{arg\,max}

\newcommand{\itxt}{\texttt{I}}

\newcommand{\review}{\texttt{R}}
\newcommand{\txt}{\texttt{T}}

\newcommand{\prior}{p}
\newcommand{\vprior}{\bm{\prior}}
\newcommand{\posterior}{q}

\newcommand{\EGPT}{\text{Elicitation}^{\text{GPT}}}
\newcommand{\wght}{w}

\newcommand{\statetxt}{\texttt{t}}

\newcommand{\SO}{O_S}

\newcommand{\QAO}{O_A}

\newcommand{\iscore}{s}

\newcommand{\vsr}{\text{ASR}}
\newcommand{\smallspac}[1]{\vspace{0mm}}

\begin{document}

\title{Aligned Textual Scoring Rules}

% It is OKAY to include author information, even for blind
% submissions: the style file will automatically remove it for you
% unless you've provided the [accepted] option to the icml2025
% package.

% List of affiliations: The first argument should be a (short)
% identifier you will use later to specify author affiliations
% Academic affiliations should list Department, University, City, Region, Country
% Industry affiliations should list Company, City, Region, Country

% You can specify symbols, otherwise they are numbered in order.
% Ideally, you should not use this facility. Affiliations will be numbered
% in order of appearance and this is the preferred way.
%\icmlsetsymbol{nt}{*}

\author{Yuxuan Lu\thanks{Peking University, Computer Science. The work is done when Yuxuan Lu is a visiting Ph.D.\ student at Northwestern University. E-mail: \texttt{yx\_lu@pku.edu.cn}.} \and Yifan Wu\thanks{Northwestern University, Computer Science. E-mail: \texttt{yifan.wu@u.northwestern.edu}.} \and Jason Hartline\thanks{Northwestern University, Computer Science Department. E-mail: \texttt{hartline@northwestern.edu}.} \and Michael J. Curry\thanks{University of Illinois Chicago, Computer Science. E-mail: \texttt{mjc@uic.edu}.}
}

\date{}

\maketitle

\begin{abstract}
Scoring rules elicit probabilistic predictions from a strategic agent by scoring the prediction against a ground truth state. A scoring rule is \emph{proper} if, from the agent's perspective, reporting the true belief maximizes the expected score. With the development of language models, \citet{WH-24} proposes a reduction from textual information elicitation to the numerical (i.e.\ probabilistic) information elicitation problem, which achieves provable properness for textual elicitation. However, not all proper scoring rules are well aligned with human preference over text. Our paper designs the Aligned Scoring rule ($\vsr$) for text by optimizing and minimizing the mean squared error between a proper scoring rule and a reference score (e.g.\ human score). Our experiments show that our $\vsr$ outperforms previous methods in aligning with human preference while maintaining properness. 
\end{abstract}
\smallspac{8}
\section{Introduction}

%Information Elicitation has become an important question when data-driven algorithms and AI systems rely crucially on the quality of input information. 

The theory of proper scoring rules is well established for elicitation of numerical information, such as the probability of a random state \citep{mcc-56, sav-71}, the mean of a distribution \citep{AF-12}, and is widely used in practice \citep{danz2022belief, hossain2013binarized, mobius2022managing}. Proper scoring rules score the quality of a probabilistic prediction by comparing to the ground truth random state. By scoring a strategic agent, proper scoring rules are mechanisms that incentivize truthful prediction. 
Information Elicitation is an important area of research that has recent practical importance due to the reliance of data-driven algorithms and AI systems on high-quality input. 

% Jason: this is better as a separate paragraph
For example, in peer grading, students report predictions about their peers' homework correctness (the random state). The instructor spot-checks homework submissions and reveals the ground truth correctness. The student's prediction is then scored in comparison to the ground truth. A scoring rule is \emph{proper} (a.k.a.\ truthful) if, from the peer's perspective, truthfully reporting her belief about the correctness maximizes expected score. 

The recent development of large language models (LLM) has enabled the evaluation of textual information. Textual reports can encode richer information than numerical predictions. For peer grading, answering open-ended review questions facilitates the students' learning process better than checking pre-specified numerical rubrics. One approach to incentivize high-quality textual review from students is to score peer reviews by querying LLM to compare student reviews with the ground truth instructor review. Moreover, studies on language-model-generated evaluation systems, i.e.,\ LLM-as-Judge \citep{zheng2023judging, fu2024gptscore}, have demonstrated that language models often align closely with human judgments when scoring text quality.

%While offering scalability, language-model-generated evaluation may not offer provable guarantees such as truthfulness (i.e.\ robustness to strategic manipulations) in sensitive applications such as peer grading in education.
Language-model-generated evaluations offer scalability but, unfortunately, lack provable guarantees such as truthfulness, leaving them vulnerable to strategic manipulation. 
%Shown in \citet{WH-24}, when scored by language models, fabricating comments may increase the expected score from submitting a peer review.
For example, when language models score peer reviews, fabricated comments may receive a higher expected score \citep{WH-24}.
To address this issue, \citet{WH-24} propose a reduction from textual elicitation problem to numerical elicitation problem. \citet{WH-24} views a language model as an oracle accepting \emph{summarization} and \emph{question-answering} queries, where summarization identifies a scoring rubric with states for elicitation, and question-answering maps text to numerical reports and states.
By implementing any numerical scoring rule over the identified space of rubrics, the scoring mechanism inherits provable properness when the language oracle is perfect and achieves adversarial robustness when the language oracle has errors. However, the scoring rules might not be aligned with preferences.

The goal of our paper is to align a provably proper textual proper scoring rule with preferences, e.g.\ human preferences. With the reduction framework in \citet{WH-24}, we optimize proper scoring rules to align with an exogenously given score that reflects a preference or a scoring rubric. For the peer grading application, we align the scoring rule to two reference scores: 1) the instructor score of peer reviews, and 2) the LLM-Judge score, by quering LLM to compare a peer review with the ground truth. While neither of these reference scores are proper, our optimization framework converts the reference scores into a proper score. %While the scoring mechanism in \citet{WH-24} maintains provable properness, there is no guarantee that every proper scoring rule applied to rubrics aligns with human preference. Following the reduction framework in \citet{WH-24}, we develop Human-Aligned Scoring Rule (\vsr) by optimizing for alignment with a reference score (e.g.\ the score given by instructor). 
%over the space of separate scoring rules. 
%A separate scoring rule applies a single-dimensional scoring rule to each summary point, and averages across single-dimensional scores. 

Our Aligned Scoring Rule ($\vsr$) is simple, provably truthful, and interpretable. We minimize the Mean Squared Error (MSE) of $\vsr$ with the reference score. We optimize over the space of separate scoring rules, which applies a single-dimensional scoring rule to each summary point and averages across single-dimensional scores. 
The hypothesis space induces a convex optimization problem with efficient algorithms. %This hypothesis space of separate scoring rules is provably truthful, interpretable, and induces a convex optimization problem.
The separate scoring rules allow us to interpret and identify the important rubric points from reference scores, by the convexity of each single-dimensional scoring rule. % is positively correlated with the importance of that summary point in scoring, making the separate scoring rule interpretable. 

We evaluate our Aligned Scoring Rule ($\vsr$) on peer grading datasets. %We optimize for alignment with two reference scores, the instructor score and the LLM score of review quality. Our $\vsr$ can be viewed as the truthful proxy of the reference scores. 
% The LLM score offers scalability, by directly querying LLM to compare the peer review against the instructor review. 
Results show that $\vsr$ fits the reference scores effectively and outperforms baselines. We first present the result of a linear regression that predicts the reference scores from $\vsr$.
%The predicted linear line is almost an identical mapping,
The regression gives almost the identity function, showing our $\vsr$ aligns identically with reference scores. Then we present the MSE and the Pearson correlation between $\vsr$ and the reference score, in comparison with baseline methods including the best constant score and the method proposed in \citet{WH-24}. Our $\vsr$ outperforms baseline methods in both metrics. Finally, we show the interpretability of $\vsr$ by a case demonstration, where $\vsr$ identifies reasonably important and non-important rubric points for scoring. 

\smallspac{2}
\paragraph{Roadmap}
\Cref{sec: prelim} introduces the preliminaries of information elicitation and proper scoring rules, including the model of numerical elicitation in \Cref{sec: prelim numerical elicitation} and textual elicitation in \Cref{sec: prelim textual elicitation}. \Cref{sec: theory} presents the reduction from textual elicitation to numerical elicitation. \Cref{sec: provable properness} provides provable guarantees to the reduction and \Cref{sec: optimization} defines our optimization problem. \Cref{sec: implementation} describes the implementation of the language oracles in reduction. \Cref{sec: empirical} presents our empirical evaluations. \Cref{sec: dataset} and \Cref{sec: reference score} introduces our dataset and baselines for comparison. \Cref{sec: experimental results} compares the alignment performance of our $\vsr$ with baselines. \Cref{fig: case study} shows an example of the optimal $\vsr$. 

\smallspac{2}
\subsection{Related Work}
\smallspac{1}

\paragraph{Textual Elicitation} 
Several recent papers design scoring mechanisms to elicit textual information from language models. \citet{kimpara2023proper} models LLM as a distribution that generates independent and identical (i.i.d.) textual samples. The paper designs a scoring rule that scores the distribution with access to samples, to incentivize a truthful report of the distribution, while our work directly scores the quality of a text.  \citet{textual-pp} designs truthful peer prediction mechanisms that score text without ground truth, by comparing the textual report of multiple peers. \citet{WH-24} designs proper scoring rules that score text with ground truth. The main goal of \citet{textual-pp, WH-24} is truthfulness (a.k.a. properness), which does not consider optimization. On the contrary, our work optimizes over the space of proper scoring rules for alignment.

\smallspac{2}

\paragraph{Grading with LLMs} Recent work studies the use of LLMs in grading textual reports from students. \citet{gg-natural-questions} studies grading via similarity between the vector embedding of the student report and ground truth. They show that the vector embedding approach works well for simple binary questions, but not for multiple-choice and more complex questions. \citet{SSNV-23} prompts a language model to compare student reports to ground truth, which is shown to have low Pearson correlation with instructor scores. Instead of directly prompting, our approach identifies scoring rubrics and optimizes for alignment while maintaining properness, thus having more favorable results. 

\smallspac{3}
\paragraph{Automated Mechanism Design and Differentiable Economics}

Automated mechanism design (AMD) is the use of computational techniques to search for good mechanisms on specific problem instances.
The earliest works in this area use linear programming~\citep{Conitzer03Automated,Conitzer03Automateda,Sandholm07Automated,Conitzer04Selfinterested};
others frame the problem in terms of learning theory, where the goal is to choose a high-performing mechanism from some class given access to samples from the type distribution~\citep{Roughgarden16Ironing,Morgenstern16Learning,Morgenstern2015Pseudo,balcan2008reducing,feldman2014combinatorial,hsu2016prices,Balcan16Sample,Balcan18General,Balcan18Dispersion}.
A body of work sometimes called ``differentiable economics'' applies the tools of modern deep learning to learn good mechanisms, either using neural networks as general function approximators~\citep{Dutting22Optimal}, or using specially-designed architectures which guarantee strategyproofness in single-agent~\citep{Shen21Automateda,Dutting22Optimal,Curry24Automated} and multi-agent settings~\citep{Curry22Differentiable,Duan23Scalablea,Wang24GemNet}.

\smallspac{1}
Like early work on AMD, the current work also solves a convex optimization problem (minimize loss subject to properness constraints), and like the learning-theoretic work, we minimize expected loss from a relatively small number of samples. Given the ability to collect or synthesize more training data, applying the flexible function approximators of differentiable economics to our setting could be a promising direction for future work. %\mjc{can probably cut this line for space if needed}

\smallspac{3}
\paragraph{Optimization of Scoring Rules}
There is an extensive literature that characterizes proper scoring rules for numerical elicitation \citep{mcc-56, sav-71}. Recently, a line of literature works on the optimization of scoring rules subject to normalization constraints such as boundedness. \citet{LHSW-22} optimizes to incentivize a binary effort in peer grading, where a peer either exerts effort to refine her posterior belief or not. As a generalization, \citet{HSLW-23} considers incentivizing a multi-dimensional effort. Our paper adopts the computation framework of the optimal scoring rule in \citet{LHSW-22}. Additionally, \citet{neyman2021binary} incentivizes sequential and discrete effort, \citet{PW-22} connects proper scoring rules to contract theory, \citet{CY-21} considers robust scoring rule design that relaxes the knowledge of the prior of the designer, and \citet{chen2023learning} designs optimal scoring rules in the online setting where the information structure and the cost of signals are unknown. 

%When the error of the language model is bounded, our human-aligned scoring rule is approximately proper. 

\smallspac{3}
\section{Preliminaries}
\label{sec: prelim}
This section introduces the preliminaries of information elicitation and scoring rules we use. 

\smallspac{3}
\subsection{Numerical Elicitation}
\label{sec: prelim numerical elicitation}

The goal of the principal (mechanism designer) is to elicit numerical reports on the quality over $\npointrubric$ explicit rubric points, represented by states $\vstate = (\state_1, \dots, \state_\npointrubric)$ where each $\state_i\in [0, 1]$. The state space is $\statespace = [0, 1]^\npointrubric$.  For example, in peer grading, the rubric consists of Statement Correctness, Proof Correctness, and Clarity. A state being $1$ means the highest quality on that rubric point. The agent holds a multi-dimensional belief $\posterior\in \Delta\left([0, 1]^n\right)$ over the $\npointrubric$ states. The principal asks the agent to report the marginal means $\vreport = (\report_1, \dots, \report_{\npointrubric})$ from the report space $\rspace = [0, 1]^\npointrubric$. 

The agent is scored by a scoring rule $\score: \rspace\times\statespace\to[0, 1]$ comparing the reported marginal means $\vreport$ and the realized state $\vstate$. A scoring rule is \emph{proper} if the expected score is maximized when the agent reports the true marginal means of the state. From the agent's subjective perspective, the scoring rule incentivizes the agent to truthfully report the believed marginal means to maximize their expected score. 

\begin{definition}[Properness]
    A scoring rule is \emph{proper} for eliciting the marginal means, if for any belief distribution $\posterior\in \Delta([0, 1]^n)$ with mean $\vmean_\posterior$, and any deviation report $\vreport\in [0, 1]^\npointrubric$, 
    \begin{equation*}
        \expect{\vstate\sim \posterior}{\score(\vmean_\posterior, \vstate)}\geq  \expect{\vstate\sim \posterior}{\score(\vreport, \vstate)}.
        \smallspac{1}
    \end{equation*}
    A scoring rule is $\epsilon$-\emph{approximately} proper if for any belief distribution $\posterior\in \Delta([0, 1]^n)$ with mean $\vmean_\posterior$, and any deviation report $\vreport\in [0, 1]^\npointrubric$, 
    \begin{equation*}
        \expect{\vstate\sim \posterior}{\score(\vmean_\posterior, \vstate)}\geq  \expect{\vstate\sim \posterior}{\score(\vreport, \vstate)} - \epsilon.
    \end{equation*}
\end{definition}

\smallspac{2}
Before reporting the belief, the agent holds a prior belief with marginal means $\vprior\in [0, 1]^n$, the empirical frequency of the ground truth in samples. The agent learns and refines the belief by receiving a signal $s\in S$ correlated with the ground truth state. The signal generation follows an information structure, a joint distribution $\Delta(\statespace\times S)$ over the state space and the signal space. Upon receiving the signal, the agent Bayesian updates to a posterior belief $\posterior\in \Delta([0, 1]^\npointrubric)$. 

\smallspac{2}

\subsection{Textual Elicitation}
\label{sec: prelim textual elicitation}

Text conveys implicit information rather than explicitly listed rubric points in numerical elicitation. Textual ground truth indicates a set of $\ndim$ summary points. The reported summary points can be represented by an $\ndim$-dimensional binary vector $\vstate = (\state_1, \dots, \state_\ndim)$, where $\state_i\in \{0, 1\}$ for each $i$. State $\state_i = 1$ or $0$ means ``agree'' or ``disagree'' on the corresponding point. For example, in a peer review of an induction homework in an algorithm class, the summary points in the textual ground truth review contain $\state_1$ the correctness of the hypothesis, $\state_2$ the base case, and $\state_3, \state_4$ two details about some particular induction step. A reported text can express uncertainty on each state, e.g. ``the base case is likely correct'' as $70\%$ probability that $\state_2 = 1$ for base case. 

In our peer grading dataset, we observe that textual reports either express a state being $0$ or $1$, or have no information. Thus, we restrict our attention to proper scoring rules with report space $\report_i = \{0, 1, \bot\}$ for each $i$. We write $\prior_i$ as the empirical frequency of $\state_i = 1$ in our dataset. \Cref{assumption: know-it-or-not} interprets an uncertain report of $\bot$ as the prior $\prior_i$. 

\begin{assumption}[Know-it-or-not]
\label{assumption: know-it-or-not}
    In the peer grading dataset, the agent's posterior belief distribution $\posterior_i$ is either $0$, $1$, or the prior $\prior_i$. %a textual report expresses uncertainty on state $\state_i$ by reporting $\bot$ or ``not applicable'', which we interpret as prior, $\report_i = \prior_i$. 
\end{assumption}

\Cref{assumption: know-it-or-not} restricts the space of proper scoring rules to scoring rules for report space $\rspace = \{0, 1, \bot\}$.

\begin{definition}[Scoring Rules for Know-it-or-not Reports]
    Given the prior distributions $\vprior$, a scoring rule $\score_{\vprior}:\{0, 1, \bot\}^\ndim\times\{0, 1, \bot\}^\ndim\to [0, 1]$ for know-it-or-not reports is proper if there exists a proper scoring rule $\score:[0, 1]^\ndim\times \{0, 1\}^\ndim\to [0, 1]$, such that 
    \begin{equation*}
        \score_\prior(\vreport, \vstate) = \score(\Tilde{\report}_{\vprior}(\vreport), \vstate), 
    \end{equation*}
    where $\Tilde{\report}_{\vprior}$ maps a report to the probabilistic belief, particularly, $\bot$ to the prior:
    \begin{equation*}
        \Tilde{\report}_{\vprior}(\report_i) = \left\{\begin{array}{cc}
        \report_i     &  \text{if }\report_i\in \{0, 1\}\\
        \prior_i     & \text{else, when }\report_i = \bot.
        \end{array}\right.
    \end{equation*}
\end{definition}

A scoring rule for multi-dimensional summary points can be defined from single-dimensional scoring rules and multi-dimensional aggregations. 

\paragraph{Single-Dimensional Scoring Rule}
We introduce the V-shaped scoring rule and the single-dimensional scoring rule for know-it-or-not reports here. 

\begin{figure}[ht]
    \centering
          \begin{tikzpicture}[scale = 0.40]

        \draw [white] (0, 0) -- (11.5, 0);
        \draw (0,0) -- (10.5, 0);
        \draw (0, 0) -- (0, 5.5);

        \draw [ultra thick] plot (0, 3.57) -- (3, 2.5);
        \draw[ultra thick] plot (3, 2.5)-- (10, 5);
        \draw (0, 3.57) -- (10, 0);
        \draw (0, 1.428) -- (10, 5);

        \draw[dotted] (0, 5) -- (10, 5);
        \draw[dotted] (0, 2.5) -- (10, 2.5);

        %        \draw (-3.2, 5) node {$\util(0)=\util(1)=1$};

%        \draw (-2.8, 2.85714) node {$\score^*(\report> \priorMean, \state)$};

%        \draw (12, 0.8) node {$\score^*(\report\leq  \priorMean, \state)$};

        \draw (-0.38, 0) node {\small $0$};
        \draw (-0.38, 5) node {\small $1$};
        \draw (-0.8, 2.5) node {\small $\sfrac{1}{2}$};
        \draw (10, -0.5) node {\small $1$};
        \draw (0, -0.5) node {$0$};
        \draw (3, -0.6) node {prior $\prior$};
        \draw (3, 0) -- (3, 0.2);
        \draw (10, 0) -- (10, 0.2);

        \draw (-1.5, 1.428) node {$\score(1, 0)$};
        \draw (-1.5, 3.57) node {$\score(0, 0)$};
        \draw (11.6, 0) node {$\score(0, 1)$};
        \draw (11.6, 5) node {$\score(1, 1)$};

        \draw (5,-1.8) node {state; belief};
        \draw  (10.5,2.5) node [rotate=90] {score};
        
      \end{tikzpicture}     
        \vspace{-3mm}
    \caption{The V-shaped scoring rule, the optimal scoring rule in \citet{LHSW-22}. The $x$ axis plots the state space, while the $y$-axis plots the score. Once fixing a report, the expected score is a linear line in both the realized state and the mean of the ground truth distribution. The line from $\score(0, 0)$ to $\score(0, 1)$ is the score for a report $\report< \prior$ below prior. From $\score(1, 0)$ to $\score(1, 1)$ is the score for a report $\report>\prior$. Reporting the prior always gets a score of $\sfrac{1}{2}$ (the dotted line). The V-shaped upper envelope of the two linear lines forms the expected score of a truthful agent. }
        \label{fig: v shape}
        \smallspac{2}
\end{figure}
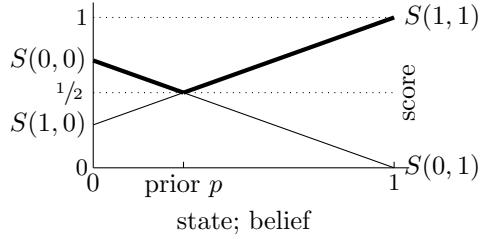

The V-shaped scoring rule is introduced in \citet{LHSW-22} as the optimal scoring rule that incentivizes a binary effort, when the agent can choose to exert effort and update her belief from prior to posterior. \citet{WH-24} tests aggregations over the V-shaped scoring rule. The V-shaped scoring rule partitions the report space into a ternary space: a report higher than the prior mean, lower than the prior mean, or the prior mean $\prior$. \Cref{fig: v shape} depicts a V-shaped scoring rule with $\prior<\frac{1}{2}$. 
\begin{definition}[V-shaped Scoring Rule]
    Given prior mean $\prior\in [0, 1]$, a V-shaped scoring rule $\score:[0, 1]\times[0, 1]\to [0, \frac{1}{2}]$  is defined by
          \begin{equation*}
      \score_{\prior} (\report, \state) = \left\{\begin{array}{cc}
      \frac{1}{2} -\frac{1}{2}\cdot \frac{\state - \prior}{1-\prior}  &  \text{if }\report < \prior\\
        \frac{1}{2} +\frac{1}{2}\cdot \frac{\state - \prior}{1-\prior}    & \text{if }\report > \prior\\
        \frac{1}{2} & \text{else}
      \end{array}
      \right.
   \end{equation*}
   When $\prior\in (\frac{1}{2}, 1]$, the score is symmetric, i.e.\ $\score_\prior(\report, \state) = \score_{1-\prior}(1-\report, 1-\state)$. 
\end{definition}

\begin{figure}[ht]
    \centering
          \begin{tikzpicture}[scale = 0.40]

        \draw [white] (0, 0) -- (11.5, 0);
        \draw (0,0) -- (10.5, 0);
        \draw (0, 0) -- (0, 5.5);

       % \draw [ultra thick] plot (0, 2.8) -- (3, 2.15);
        %\draw[ultra thick] plot (3, 2.5)-- (10, 5);
        \draw (0, 2.8) -- (10, 0.5);
        
        \draw (0, 1.) -- (10, 5);

        \draw[dotted] (0, 5) -- (10, 5);
        \draw (0, 2.2) -- (10, 3.8);

        %        \draw (-3.2, 5) node {$\util(0)=\util(1)=1$};

%        \draw (-2.8, 2.85714) node {$\score^*(\report> \priorMean, \state)$};

%        \draw (12, 0.8) node {$\score^*(\report\leq  \priorMean, \state)$};

        \draw (-0.38, 0) node {\small $0$};
        \draw (-0.38, 5) node {\small $1$};
        %\draw (-0.8, 2.5) node {\small $\sfrac{1}{2}$};
        \draw (10, -0.5) node {\small $1$};
        \draw (0, -0.5) node {$0$};
        \draw (3, -0.6) node {prior $\prior$};
        \draw (3, 0) -- (3, 0.2);
        \draw (10, 0) -- (10, 0.2);

        \draw (-1.5, 0.8) node {\small$\score(1, 0)$};
        \draw (-1.5, 2) node {\small$\score(\bot, 0)$};
        \draw (-1.5, 3) node {\small$\score(0, 0)$};
        \draw (11.6, 0.4) node {\small$\score(0, 1)$};
        \draw (11.6, 5) node {\small$\score(1, 1)$};

         \draw (11.6, 3.8) node {\small$\score(\bot, 1)$};

        \draw (5,-1.8) node {state; belief};
        \draw  (10.5,2.5) node [rotate=90] {\small score};
        
      \end{tikzpicture}     
    \vspace{-3mm}
    \caption{An example of a single-dimensional scoring rule for know-it-or-not reports. Each report in the ternary space corresponds to a linear line. The scoring rule can be depicted by three linear lines. Properness requires that, when the belief (or, equivalently, the ground truth) is $\report$, the line with the highest expected score is on the line corresponding to report $\report$. }
    \label{fig: single d}
    \smallspac{2}
\end{figure}
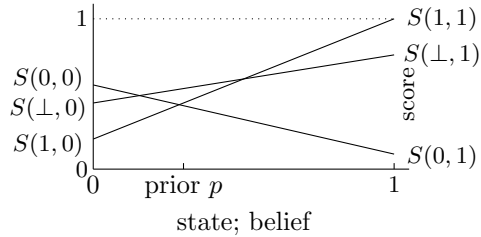

A single-dimensional scoring rule for know-it-or-not reports can be characterized by nine values: $\score(\report, \state)$ for $\report\in \{0, 1, \bot\}$ and $\state\in \{0, 1\}$. The definition of properness simplifies to \Cref{def: know it or not proper}. A V-shaped scoring rule is a special case of a single-dimensional scoring rule for know-it-or-not reports, where the score of reporting $\bot$ is fixed at $\frac{1}{2}$. \Cref{fig: single d} presents a graphical illustration of such a scoring rule.
\begin{definition}
\label{def: know it or not proper}
    Given prior $\prior$, a single-dimensional scoring rule for know-it-or-not reports is proper if 
\begin{align*}
    \score(\state, \state)\geq \score(\report, \state), \quad\quad \quad\,\forall \state\in \{0, 1\}, \forall \report\in \{0, 1, \bot\}\\
    \expect{\state\sim \prior}{\score(\bot, \state)}\geq \expect{\state\sim \prior}{\score(\report, \state)}, \qquad\forall \report\in \{0, 1, \bot\}
\end{align*}
\end{definition}

\paragraph{Multi-Dimensional Aggregations}
A multi-dimensional aggregation operates over single dimensional scoring rules and preserves properness. 

\begin{definition}
    Given single dimensional scoring rules $\score_1, \dots, \score_\ndim$ where each $\score_i:[0, 1]\times[0, 1]\to [0, 1]$, a multi-dimensional scoring rule $\score:[0, 1]^m\times[0, 1]^m\to [0, 1]$ is aggregated from $\score_1, \dots, \score_\ndim$ if 1) $\score$ is proper, and 2) there exists aggregation function $A$ such that 
       \begin{equation*}
        \score(\report_1, \dots, \report_\npointrubric; \cdot) = A\big(\score_1(\report_1; \cdot), \dots, \score_\npointrubric(\report_\npointrubric; \cdot)\big).
    \end{equation*}
\end{definition}

We introduce two aggregations, the separate aggregation and the max-over-separate (M) aggregation.

We optimize over the space of separate scoring rules \citep{LHSW-22}. \citet{WH-24} also tests the averaged V-shaped scoring rule (AV). 
\begin{definition}
    Given scoring rules $\score_1, \dots, \score_\ndim$, a separate scoring rule  is the weighed average $\score = \sum_{i\in[\ndim]}\wght_i\score_i$, with weights $\wght_1, \dots, \wght_\ndim$ such that $\sum_{i\in [\ndim]}\wght_i = 1$. 
\end{definition}

The max-over-separate scoring rule scores an agent by the dimension on which the agent has the highest expected score. It can be implemented by asking the agent to pick her favorite dimension and score on that dimension. \citet{WH-24} tests the max-over-separate V-shaped scoring rule (MV), the optimal scoring rule in the multi-dimensional report. We will compare our Aligned Scoring Rule with the MV scoring rule. 
\begin{definition}
    [Max-Over-Separate]
        Given scoring rules $\score_1, \dots, \score_\ndim$, a max-over-separate scoring rule  is 
        \begin{equation*}
            \score(\vreport, \vstate) = \score_i(\report_i, \state_i), \text{ where }i = \argmax_{i'}\expect{\state_{i'}}{\score_{i'}(\report_{i'}, \state_{i'})}.
        \end{equation*}
\end{definition}

\section{Aligned Scoring Rule: Algorithm}
\label{sec: theory}

In this section, we present our design of Aligned Scoring Rule ($\vsr$), which reduces textual elicitation to numerical elicitation and optimizes for human alignment in peer grading. \Cref{sec: provable properness} list the provable properness guarantees of the reduction from \citet{WH-24}. \Cref{sec: optimization} describes our optimization method for alignment. 

Following \citet{WH-24}, we model the language model as an oracle accepting \emph{Summarization} and \emph{Question-Answering} queries, which are fundamental natural language processing tasks \citep{bar2020arguments, clark2019boolq, rajpurkar-etal-2016-squad}. The Summarization oracle outputs a list of summary points from a list of texts. The Question-Answering oracle identifies whether a text agrees or disagrees with a summary point. 
\smallspac{2}
\begin{description}
    \item[Summarization] $\SO$, summarizes a list of textual report into summary points. 
    \begin{description}
        \item [Input] A list of texts $\txt_1, \dots, \txt_N$.
        \item[Output] A list $[\statetxt_1, \ldots, \statetxt_\ndim]$ of all summary points from texts.
    \end{description}
    \smallspac{1}

    \item[Question-Answering] $\QAO$ determines whether a text agrees or disagrees with a summary point, or is not applicable. 
    \begin{description}
        \item[Input] One text $\txt$ and a summary point $\statetxt$. 
        \item [Output] Output ``disagree'' $0$, ``agree'' $1$, or ``NA'' $\bot$. % in this dimension based on textual report $r_\txt$, where $\report=0$ indicates support for the negative judgemnt, $\report=\bot$ indicates not applicable, and $\report=1$ indicates support for the positive judgemnt.
    \end{description}
\end{description}

\smallspac{2}
We describe $\EGPT$ from \citet{WH-24} here. Following \Cref{assumption: know-it-or-not}, we map a report $\bot$ to the prior report, the empirical frequency of a summary point. The clustered nature of the peer grading application enables the identification of the empirical frequency. The dataset is partitioned in advance into clusters. Each cluster contains $N$ peer grading tasks, where the homework submission are all from the same assignment, thus applicable to the same set of grading rubrics. 

\paragraph{Input}
\smallspac{2}
\begin{itemize}
    \item A cluster of $N$ ground truth reviews $\{\itxt_1, \dots, \itxt_N\}$ on submissions to the same homework assignment. 

        \item One reported review $\review_k$ on the $k$th submission. 

    \item A proper scoring rule $\score$ for know-it-or-know beliefs. 
\end{itemize}

We will write the identified states and reports by the language oracle as $\hat{\vstate}$ and $\hat{\vreport}$, respectively. 

\paragraph{Algorithm ($\EGPT$)}
\smallspac{2}
\begin{itemize}
    \item (Summarization) Summarize instructor reviews into summary points. 

    $\{\statetxt_1, \ldots, \statetxt_\ndim\} = \SO(\{\itxt_i\}_{i\in [N]})$.
    \item (Question-Answering) Map truth $\itxt_i$ to state space.

    For each instructor review $j\in [N]$ and each summary point $i\in [\ndim]$, $\state_i^{j} = \QAO(\itxt_j, \statetxt_i)$.

    Calculate the prior of each state $\prior_i = \frac1N{\sum_{j}\state_i^j}$. 

    \item (Question-Answering) Map the review to report space. 

    For each summary point $i\in [\ndim]$, $\hat{\report}_i = \QAO(\review_k, \statetxt_i)$. 

    \item Apply proper scoring rule for know-it-or-not reports. 

    Output $\score_{\vprior}(\vreport, \vstate^k)$\footnote{Note that the ground truth may have $\bot$ in our implementation. In such a case, we score the student by the expected score where the binary state is drawn from the prior. }.
\end{itemize}

\subsection{Provable Properness}
\label{sec: provable properness}

We list the provable property of the reduction here, including the case that the language oracle makes errors and adversarial robustness. 

The correctness of summarization $\SO$ does not affect the truthfulness of $\EGPT$. To see this, even when $\SO$ identifies the wrong summary points, $\EGPT$ is still proper as long as $\QAO$ correctly identifies the numerical states and reports corresponding to the summaries. We assume $\QAO$ is perfect on the ground truth side, as the ground truth reviews often clearly state opinions on summary points. 

% When the language oracle $\QAO$ makes bounded errors on the ground truth side, $\EGPT$ is approximately truthful. 

% \begin{theorem}[\citealt{WH-24}]
%     If the scoring rule $\score$ is a separate scoring rule, and if $\QAO$ has error 
%     \begin{equation}
%         \expect{\hat{\state_i}}{|\hat{\state_i} - \state_i|\mid \itxt}\leq \epsilon, \forall i, \forall \itxt,
%     \end{equation}
%  $\EGPT$ is $2\epsilon$-approximately proper. 
% \end{theorem}

When the language oracle $\QAO$ is non-inverting on the report side, $\EGPT$ is proper. 

\begin{definition}
    [Non-inverting $\QAO$]
The question-answering oracle for know-it-or-not beliefs is non-inverting if the probability of inverting the report is strictly less than $\frac{1}{2}$, i.e.\ $\Pr[\hat{\report}_i\neq \report_i | \review]\leq \frac{1}{2}$ for any $i$ and any $\review$.
\end{definition}

\begin{theorem}[\citealt{WH-24}]
\label{corollary: proper for peer grading}
If the question-answering oracle for know-it-or-not beliefs is non-inverting for reports, $\EGPT$ is proper.
\end{theorem}

Without any assumptions on the error of the language oracle, the reduction above has adversarial robustness. 
\begin{theorem}[\citealt{WH-24}]
    If the agent has no information, the highest expected score she can achieve is at most by saying $\bot$ (i.e.\ ``I don't know''). 
\end{theorem}

\subsection{Optimization for Alignment}
\label{sec: optimization}

While $\EGPT$ presents a framework for reducing textual elicitation to numerical elicitation, not all proper scoring rules align well with the instructor preferences. Thus, our Aligned Scoring Rule ($\vsr$) optimizes over a space of separate scoring rules and selects the one that aligns best with the reference score, i.e., the instructor score of a peer review. Our optimization framework follows the computation of optimal scoring rule in \citet{LHSW-22}. Our Aligned scoring rule can be viewed as a truthful proxy of the instructor score. 

Fixing summary points $\{\statetxt_1, \ldots, \statetxt_\ndim\}$ and prior $\vprior$, our optimization objective minimizes the mean squared error (MSE) between $\EGPT$ score and the reference score (e.g.\ instructor score). Our optimization problem is shown in 
Program \ref{eq: opt general} with $\iscore$ normalized to $[0, 1]$. 
\begin{align}
\label{eq: opt general}
    \min_{\{\score\}_{i\in [\ndim]}}\ \  &\expect{(\vreport, \vstate, s)}{\left(\score(\vreport, \vstate) - \iscore\right)^2}\\
    \text{s.t.} \ \ & \score \text{ is proper}\nonumber\\
   &\score(\cdot, \cdot)\in [0, 1]\nonumber
\end{align}

%instructor score
We optimize over the space of separate scoring rules, the sum of single-dimensional proper scoring rules $\{\score_i\}_{i\in [\ndim]}$ for know-it-or-not reports. A separate scoring rule is simple and interpretable, where the convexity of single-dimensional scores can identify the importance of each dimension. We present a case study of the interpretability in \Cref{sec: case demonstration}. Program \ref{eq: opt} shows the simplified optimization problem for separate scoring rules. The properness constraint follows properness for know-it-or-not reports in \Cref{def: know it or not proper}.
\begin{align}
\label{eq: opt}
    \min_{\{\score_i\}_{i\in [\ndim]}}\ \  &\expect{(\vreport, \vstate, s)}{\left(\sum_{i\in [\ndim]}\score_i(\report_i, \state_i) - \iscore\right)^2}\\
    \text{s.t.} \ \ & \text{for any dimension } i, \tag{Properness}\nonumber\\
    &\quad \text{for any }\report_i\in \{0, 1, \bot\}\nonumber\\
    &\qquad \score_i(\state_i, \state_i)\geq \score_i(\report_i, \state_i), \forall \state_i\in \{0, 1\}\nonumber\\
    &\qquad \expect{\state_i\sim \prior_i}{\score_i(\bot, \state_i)}\geq \expect{\state_i\sim \prior_i}{\score_i(\report_i, \state_i)}\nonumber\\
   &\sum\nolimits_{i\in [\ndim]}\score_i(\report_i, \state_i)\in [0, 1], \forall \vreport, \vstate \tag{Boundedness}\nonumber
\end{align}

 %says, on dimension $i$, when the agent's belief is deterministically $\state_i$, reporting the state $\state_i$ gains a higher score. 
%The second properness constraint says, when the agent's belief follows prior $\prior_i$, i.e.\ she does not have information on that dimension, reporting $\bot$ achieves a higher expected score. 

Our optimization problem with separate scoring rules is convex. Note that the same formulation may not be convex for other spaces of multi-dimensional scoring rules, e.g.\ max-over-separate scoring rules. 

\begin{corollary}
\label{corollary: convex opt}
    Optimization problem \ref{eq: opt} is convex. 
\end{corollary}
To see \Cref{corollary: convex opt}, note that for each dimension, we have six variables: $\score_i(\report_i, \state_i)$ for $\report_i\in \{0, 1, \bot\}$ and $\state_i\in \{0, 1\}$. Both our objective and constraints are convex in the variables. Since optimization problem \ref{eq: opt} is convex, we optimize with the gradient descent algorithm over samples.

\section{Implementation of Language Oracles}
\label{sec: implementation}

We describe our implementation of the language oracle here. 

\subsection{Summarization Oracle}

The implementation of the summarization oracle includes three steps: summarizing instructor reviews, preparing negative/positive statement pairs from reviews, and clustering negative/positive statement pairs. Note that instead of directly clustering summary statements by similar meanings, for each statement from the reviews, we concatenate the statement with another of the opposite meaning to prepare a pair of negative/positive statements. The negative/positive statement pairs improve the robustness of LLM clustering. When each summary point consists of negative/positive statement pairs, the semantic meaning of each state can be viewed as neutral, avoiding opposite statements being identified as different states for elicitation. 

\smallspac{2}
\paragraph{Input} A list of $N$ instructor reviews $[\itxt_1, \dots, \itxt_N]$.
\paragraph{Output} A list $[\statetxt_j]_{j\in m}$ of summary points from reviews. 
\smallspac{2}
\paragraph{Implementation}
    We provide a toy prompt with each step below. The real prompts we use are listed in \Cref{app:implement}.
    \smallspac{2}
    \begin{itemize}
        \item Summarize each instructor review into summary points. %Write the set of statements from all reviews as $\{\hat{\statetxt}_j\}$
        %$\itxt_i$ 
\smallspac{2}
        
        \textit{\textbf{Toy prompt}: Carefully read the entire review comment. Extract all evaluative statements from the review. These should be comments that assess the quality, strengths, weaknesses, and suggestions. Ignore purely descriptive or meaningless statements. Ignore statements purely about specific scores and ratings. Create an indexed list of these evaluative statements.}
        \smallspac{1}
        \item Transform each statement into negative/positive pairs.
        %$\hat{\statetxt}_{j\pm}$

        \smallspac{1}

        \textit{\textbf{Toy prompt}: You are tasked with creating opposite evaluative statements for a given list of evaluative statements. For each statement provided, you need to create a new statement that has the same content but expresses the opposite emotion or sentiment.}

        \item Cluster the negative/positive pairs of summary points. The semantic meaning of each cluster is identified as the dimension for elicitation, $[\statetxt_j]_{j\in [\ndim]}$.

        \textit{\textbf{Toy prompt}: You will be given a list of opinion pairs, where each pair consists of a positive opinion and its corresponding negative opinion. Your task is to analyze these pairs and cluster them based on similarity.}
    \end{itemize}

\subsection{Question-Answering Oracle}

We directly query LLM to identify whether a review $\review$ is positive or negative for a summary point $\statetxt$. %The Question-Answering Oracle takes as input a review and a summary point (including the positive and negative opinion), and then outputs whether the review supports the positive one, the negative one, or no relevant content.

\smallspac{2}
\paragraph{Input} One review $\review$ and a summary point $\statetxt$.
\smallspac{2}
\paragraph{Output} Positive ($1$), negative ($0$), or NA ($\bot$).

\textbf{Implementation} We provide an toy prompt below. The real prompt we use are listed in \Cref{app:implement}.%We query LLM to identify the meaning of review $\review$ on a cluster $\statetxt$ of statement pairs of the same meaning. 

\textit{\textbf{Toy prompt}: Your task is to infer which of the given positive/negative opinions is correct based on the provided review comment. For each opinion pair, carefully read and understand both the positive and negative opinions. Conclude whether the review supports the positive, the negative, or neither opinion.}

\smallspac{2}
\section{Empirical Evaluation}
\label{sec: empirical}
\smallspac{2}

We describe our dataset and evaluation metric in \Cref{sec: dataset}, our reference scores used for alignment in \Cref{sec: reference score}, and our experimental results in \Cref{sec: experimental results}. We depict the Aligned Scoring Rule ($\vsr$) for one example homework assignment in \Cref{sec: case demonstration}.
\smallspac{2}

\subsection{Dataset and Evaluation Metric}
\label{sec: dataset}
\paragraph{Dataset} We present results from peer grading data in two undergraduate algorithm classes. Our dataset includes 22 assignments in total.\footnote{Algorithm Class 1: 276 reviews by 23 peers on 89 submissions across 12 assignments. Algorithm Class 2: 240 reviews by 24 peers on 59 submissions across 10 assignments.} Each assignment has $6$ to $8$ homework submissions. Each homework submission has one instructor review (i.e.\ ground truth) and $6$ to $8$ peer reviews. Each peer review has an instructor score in $[0, 10]$. %Both peer reviews and instructor reviews provide comments on three explicit rubric points: Answer/Algorithm, Proof/Analysis, and Clarity.
\smallspac{2}
\paragraph{Metric} We report the \textit{Mean Squared Error}, the \textit{Pearson correlation coefficient}, and the \textit{Spearman rank correlation coefficient} of our $\vsr$ compared with reference scores. 
\begin{itemize}
    \item MSE quantifies the average magnitude of prediction errors.
    \item Pearson correlation assesses the strength of the linear relationship between predicted scores and reference scores, capturing whether the model correctly preserves the relative ordering. A Pearson correlation is in $-1$ to $1$, where $0$ means no correlation, $1$ means perfectly correlated, and $-1$ means perfectly negatively correlated. A Pearson correlation $>0.4$ is thought as a moderate correlation, $>0.6$ a strong correlation, and $>0.8$ very strong and almost linear. 
    \item Spearman rank correlation assesses the correlation between two ranks. Similar as Pearson correlation, the Spearman rank correlation is in $[-1, 1]$, where $0$ means the two ranks are not correlated, $1$ means identical ranking, and $-1$ means reversed ranking. The assessment of values are also the same as above. 
\end{itemize} 

%We also run the same experiment on data from \yw{Course xxxx}, with the results provided in the Appendix. %We only include students who completed reviews for all assignments.

%, from both peers and instructors, includes scores and comments on three rubric points:  Instructors also rate the quality of each peer's review (both numerical and textual) on a $[0, 10]$ scale. We only include students who completed reviews for all assignments.

%Since these two sets of data come from the same course offered in two consecutive years, we merge them when reporting the results.

% \paragraph{Evaluation Metric}  

%\yw{How are these calculated}
\smallspac{2}
\subsection{Reference Score}
\label{sec: reference score}

%In the peer grading case, each peer review request of a submission is a prediction task. The instructor review, denoted as $\state_\txt$, serves as the textual ground truth. Tne peer review, $\report_\txt$, constitutes the textual report from an agent. Furthermore, we define the \emph{reference reward} for the agent's report in two different ways:
We optimize for alignment with two reference scores, the Instructor Score and the LLM-Judge Score. 
\smallspac{2}
\begin{description}
    \item[Instructor Score] Instructor score (i.e., human preference) from our dataset. %The instructor scores the quality of a peer review based on their own instructor review and the peer review. This data was collected as part of our dataset.
    \smallspac{2}
    \item[LLM-Judge Score] We query a language model to grade the peer review against the instructor review based on a given peer review scoring rubric.
    %\yuxuan{robustness check in app}%, e.g.\ $6/10$ for a slightly flawed peer review, $8/10$ for a correct review, $10/10$ for an extraordinarily good review, etc. 
    \smallspac{3}
\end{description}

\begin{figure}[h]
\centering
\includegraphics[width=.75\linewidth]{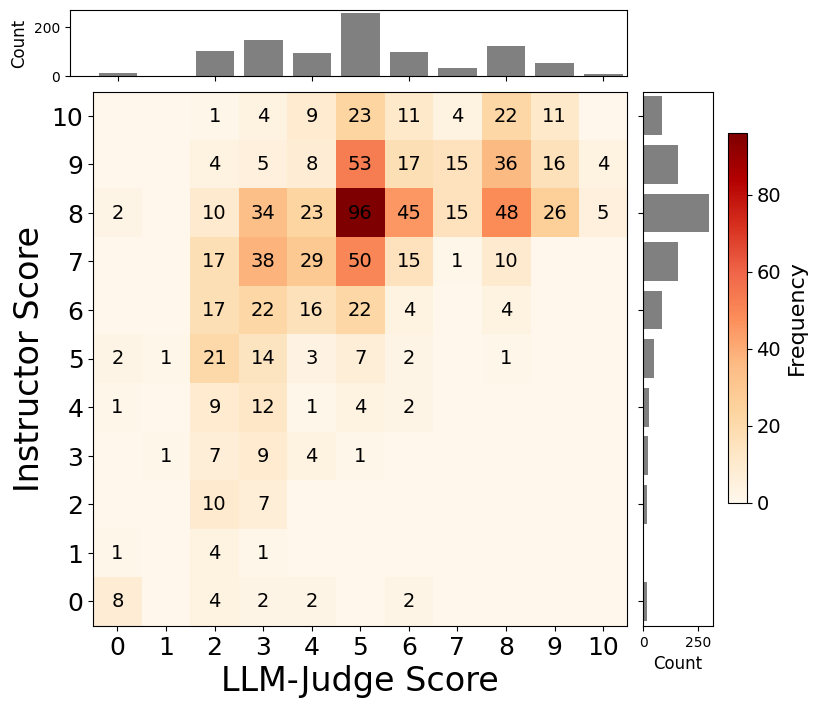}
\smallspac{3}
\caption{Joint distribution (instructor score vs. LLM-Judge score)}
\label{fig:correlation}
\smallspac{3}
\end{figure}

There is a high correlation between the Instructor Score and LLM-Judge score. \Cref{fig:correlation} presents the empirical joint distribution of Instructor Score and LLM-Judge Score for all data, with a Pearson correlation of 0.5540. %, which shows there is a relatively high correlation between instructor and LLM reference score.
The results show that LLM-Judge score can serve as a substitute for the costly and noisy instructor score, improving the scalability and the robustness of the peer grading system, which is consistent with previous studies of the LLM-as-Judge method, e.g., \citealp{zheng2023judging, hackl2023gpt}, etc.  %While human scores are costly and noisy, we calculate the average output of LLM-Judge scores over multiple generations, which improves the score consistency. %allow noise reduction through averaging over multiple runs.

Note, the instructor and LLM-judge reference scores are not proper and therefore might encourage peer reviewers to engage in strategic behavior like guessing or adding irrelevant statements \citep{WH-24}.
%See \citet{WH-24} for manipulation strategies for LLM-judge score.
Our method of aligning a proper scoring rule to these references can be viewed as converting these non-proper scores into proper ones.

%We investigate both Instructor-as-reference and LLM-as-reference because, if the reference rewards provided by LLMs are highly correlated with those from instructors, LLMs can serve as a practical substitute for scoring peer reviews. This substitution significantly improves scalability, as it eliminates the need for instructors to evaluate every peer review and limits their involvement to scoring submissions only.

\smallspac{2}
\subsection{Experimental Results}
\label{sec: experimental results}
%This section details the experimental results, focusing on the outcomes and findings from two algorithm courses. 
%We experiment with \textsf{gemini-2.5-flash-preview-04-17} except for clustering. For clustering, we employed \textsf{gemini-2.5-pro-preview-05-06} due to its proficiency in handling long contexts. We set the temperature to $0$ with the ``thinking'' feature disabled.
We present our experimental results in this section. First, we show that a linear regression fitting the reference score from our $\vsr$ results in a nearly-identity linear fit. We then present the MSE and the Pearson correlation coefficients and compare with baselines. We use the \textsf{gemini-2.5} series models for the LLM-Judge and the LLM oracles in our experiments. Parameters and prompt details can be found in \Cref{app:implement}. We also tested the performance of \textsf{GPT-4.1} as the LLM-Judge on the same prompt, with the results detailed in \Cref{app:addi}.

\begin{figure}[t]
\smallspac{2}
  \centering
  \subfigure[Instructor score vs.\ $\vsr$ aligned with instructor score.]{
    \includegraphics[scale = 0.3]{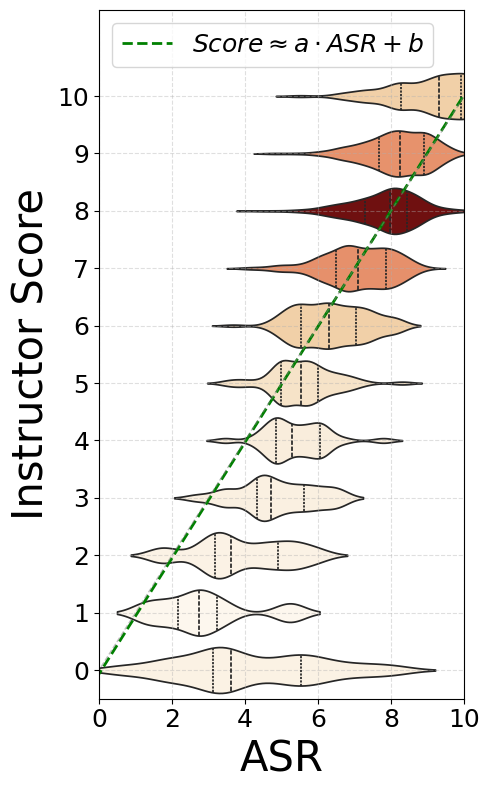}
    \label{fig:joint-inst}
  }
  \quad
  \subfigure[LLM-Judge score vs.\ $\vsr$ aligned with LLM-Judge score.]{
    \includegraphics[scale = 0.3]{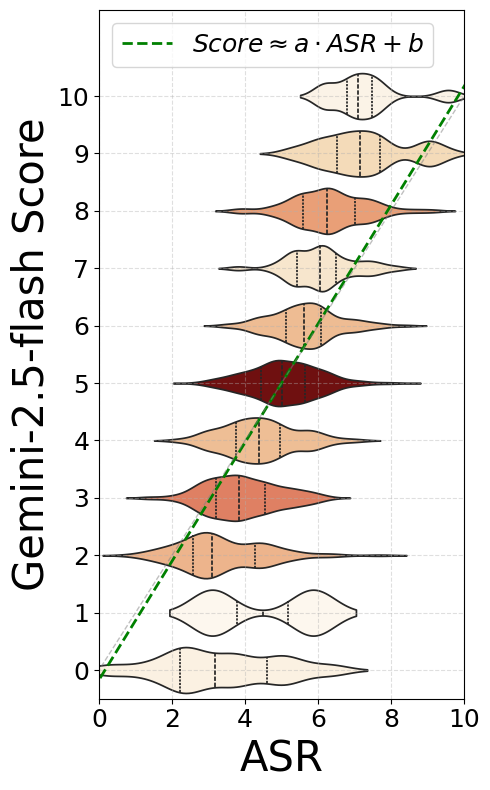}
    \label{fig:joint-llm}
  }
  \caption{Reference Scores vs. \vsr: The green dotted line represents the linear regression fitting reference score from $\vsr$. On both plots, the linear relationship is almost the identity function. }
  \label{fig:joint}
  \smallspac{3}
\end{figure}

\paragraph{Nearly-Identity Linear Fit}
The first criterion to evaluate the effectiveness of our approach is to examine whether our $\vsr$ can effectively fit the original reference scores. \Cref{fig:joint} illustrates the joint empirical distribution of the $\vsr$ scores and the reference scores, %We expect the $\vsr$ to serve as a good predictor of the original reference; 
with a regression line predicting the reference score $s$ from the $\vsr$ score $\score$. The parameters of linear regression align closely with $s = \score$. %\yw{What is this R-squared val}, together with the high R-squared value, confirms this. %Using the Instructor and LLM as references, the Pearson correlations between the reference score and the fitted sum-VSR are \yuxuan{to be added} and  respectively.

% \begin{figure}[ht]
%   \centering
%   \begin{subfigure}{\linewidth}
%     \includegraphics[width=\linewidth]{21.png}
%     \caption{Subfigure A}
%     \label{fig:subfig-a}
%   \end{subfigure}
%   \hfill
%   \begin{subfigure}{\linewidth}
%     \includegraphics[width=\linewidth]{22.png}
%     \caption{Subfigure B}
%     \label{fig:subfig-b}
%   \end{subfigure}
%   \caption{Overall caption for the figure.}
%   \label{fig:overall-fig}
% \end{figure}

\smallspac{2}
\paragraph{Comparison with Baselines}

Our Aligned Scoring Rule is compared against the following two baselines which are all truthful:
\smallspac{2}
\begin{enumerate}
    \item \textbf{Best Constant Score ($S_{\text{const}}$)}. This method outputs the best constant score for all reviews, which is the mean of the reference scores $s$ in the training data $D$. The constant score is weakly truthful.
    \[ S_{\text{const}}(\report_\txt,\state_\txt) = \sum\nolimits_{(\report, \state, s) \in D} s/|D|. 
    \smallspac{2}\]

    % \item \textbf{Simple Linear Regression}: This approach computes a linear regression based on the L2 norm $\lVert\hat{r}-\hat{\theta}\rVert_2$ between projected reports $\hat{r}$ and projected ground truths $\hat{\theta}$ from dataset $D$, and the reference reward $s$. The regression coefficient is constrained to be non-negative to ensure truthfulness. The output of this linear regression serves as the truthful scoring rule.
\smallspac{2}
    \item \textbf{Non-aligned ElicitationGPT (EGPT)}. We compare with the $\EGPT$ in \citet{WH-24}, which is not aligned to a reference, particularly, the averaged V-shaped scoring rule (AV) and the max-over-separate V-shaped scoring rule (MV). In \citet{WH-24}, the AV scoring rule is shown to align the best with instructor score. Note that the max-over-separate scoring rule is not in our hypothesis space of separate scoring rules, and does not induce a convex optimization problem. \footnote{We evaluate Spearman correlation differently from \citet{WH-24}. \citet{WH-24} evaluate the ranking of the same student's averaged scores over all peer reviews in a class, becausre the $\EGPT$ scores are not optimized and aligned in the same scale as reference scores. We evaluate each individual peer review's ranking, as our score is aligned.  } %\citet{WH-24} evaluates scoring rules adopted from \citet{LHSW-22}. We compare with the averaged V-shaped scoring rules, which are shown to align the best with instructor score in \citet{WH-24}. 
\end{enumerate}

The performance of scores is evaluated along three metrics: MSE, the Pearson correlation coefficient, and the Spearman rank correlation coefficient. Our $\vsr$ aligns best with the reference on all metrics.

\begin{table}[ht]
  \centering
  \renewcommand{\arraystretch}{.85} % Reduce row height
  \setlength{\tabcolsep}{2pt}      % Reduce column separation

  \subfigure[Reference: Instructor Score]{%
    \centering
    \begin{tabular}{lccc}
      \hline
      Method    & Squared Loss & Pearson Corr & Spearman Corr \\
      \hline
      $\vsr$      &  1.730       & 0.717        & 0.622         \\
      Constant  &  3.741       & N/A          & N/A           \\
      EGPT (AV) &  9.541       & 0.294        & 0.301         \\
      EGPT (MV) & 18.360       & 0.213        & 0.207         \\
      \hline
    \end{tabular}
  }%
  \hfill
  \subfigure[Reference: LLM-Judge Score]{%
    \centering
    \begin{tabular}{lccc}
      \hline
      Method    & Squared Loss & Pearson Corr & Spearman Corr \\
      \hline
      $\vsr$       &  2.003       & 0.705        & 0.658         \\
      Constant  &  4.136       & N/A          & N/A           \\
      EGPT (AV) &  7.053       & 0.328        & 0.338         \\
      EGPT (MV) & 17.069       & 0.246        & 0.226         \\
      \hline
    \end{tabular}
  }

  \caption{Comparison with baselines. }
  \label{tab:baseline_comparison}
\end{table}

%\yuxuan{HASR trained on human/LLM-Judge score}

\subsection{Case Demonstration}
\label{sec: case demonstration}

\begin{figure}[htbp]
\smallspac{2}
\centering
\includegraphics[width=\linewidth]{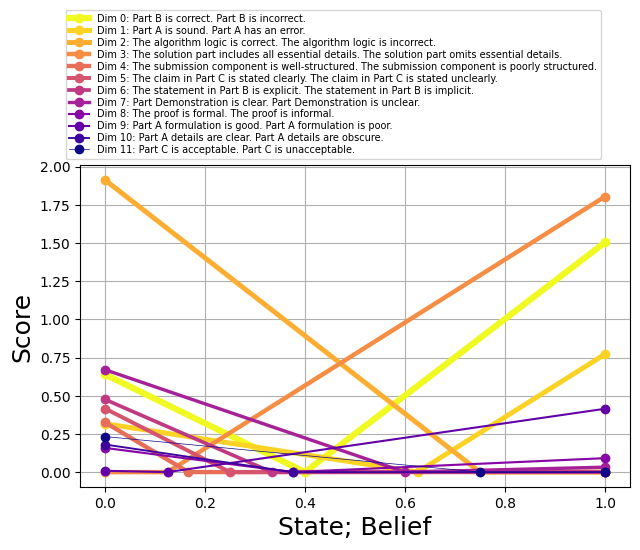}
\smallspac{3}
\caption{The visualization of $\vsr$ on one assignment in the algorithm class using instructor score as the reference. The score of $\report = \bot$ for each dimension has been shifted to zero. }%It can be observed that dimensions generally considered more important indeed have larger weights.}
\label{fig: case study}
\end{figure}

We present an example of $\vsr$ in this section. \Cref{fig: case study} visualizes the single-dimensional scoring rules. The homework assignment is on asymptotic analysis and is divided into three parts $A, B, C$, each corresponding to the asymptotic relationship between two functions. For each dimension, we plot the V-shape scoring rule for this dimension.% corresponding to reports $\report\in \{0, 1\}$. %\mjc{I find this plot very hard to read, but i don't have better suggestions for visualizing it}

From the plot, we can observe the dimensions that are not important for scoring, where the scoring line is almost linear, meaning the score does not depend on the report but only on the state. For example, we observe that the dimensions for clarity are less important, e.g., ``part A details are clear'' and ``submission well-structured''. 

We also identify important dimensions, where the two linear scoring lines form a more strongly convex function. We observe that summary points on details related to overall correctness are more important, e.g., ``Algorithm logic is correct'', ``solution omits details'', Dim 4 ``Part B is correct'', and ``Part A is sound''. 

In general, we observe that our ASR when learning Instructor Score assign more convex V-shape scoring rule to the content that is commonly considered to be more important.

%\yuxuan{recall fig 1. Rearrange order of dims}
%\yuxuan{y axis in log scale}

\newpage
\bibliography{ref}

\begin{thebibliography}{42}
\providecommand{\natexlab}[1]{#1}
\providecommand{\url}[1]{\texttt{#1}}
\expandafter\ifx\csname urlstyle\endcsname\relax
  \providecommand{\doi}[1]{doi: #1}\else
  \providecommand{\doi}{doi: \begingroup \urlstyle{rm}\Url}\fi

\bibitem[Abernethy \& Frongillo(2012)Abernethy and Frongillo]{AF-12}
Abernethy, J.~D. and Frongillo, R.~M.
\newblock A characterization of scoring rules for linear properties.
\newblock In \emph{Conference on Learning Theory}, pp.\  27--1, 2012.

\bibitem[Balcan et~al.(2008)Balcan, Blum, Hartline, and
  Mansour]{balcan2008reducing}
Balcan, M.-F., Blum, A., Hartline, J.~D., and Mansour, Y.
\newblock Reducing mechanism design to algorithm design via machine learning.
\newblock \emph{Journal of Computer and System Sciences}, 74\penalty0
  (8):\penalty0 1245--1270, 2008.

\bibitem[Balcan et~al.(2018{\natexlab{a}})Balcan, Dick, and
  Vitercik]{Balcan18Dispersion}
Balcan, M.-F., Dick, T., and Vitercik, E.
\newblock Dispersion for {{Data-Driven Algorithm Design}}, {{Online Learning}},
  and {{Private Optimization}}.
\newblock In \emph{2018 {{IEEE}} 59th {{Annual Symposium}} on {{Foundations}}
  of {{Computer Science}} ({{FOCS}})}, pp.\  603--614, October
  2018{\natexlab{a}}.
\newblock \doi{10.1109/FOCS.2018.00064}.

\bibitem[Balcan et~al.(2018{\natexlab{b}})Balcan, Sandholm, and
  Vitercik]{Balcan18General}
Balcan, M.-F., Sandholm, T., and Vitercik, E.
\newblock A {{General Theory}} of {{Sample Complexity}} for {{Multi-Item Profit
  Maximization}}.
\newblock In \emph{Proceedings of the 2018 {{ACM Conference}} on {{Economics}}
  and {{Computation}}}, pp.\  173--174, Ithaca NY USA, June 2018{\natexlab{b}}.
  ACM.
\newblock ISBN 978-1-4503-5829-3.
\newblock \doi{10.1145/3219166.3219217}.

\bibitem[Balcan et~al.(2016)Balcan, Sandholm, and Vitercik]{Balcan16Sample}
Balcan, M.-F.~F., Sandholm, T., and Vitercik, E.
\newblock Sample complexity of automated mechanism design.
\newblock \emph{Advances in Neural Information Processing Systems}, 29, 2016.

\bibitem[Bar-Haim et~al.(2020)Bar-Haim, Eden, Friedman, Kantor, Lahav, and
  Slonim]{bar2020arguments}
Bar-Haim, R., Eden, L., Friedman, R., Kantor, Y., Lahav, D., and Slonim, N.
\newblock From arguments to key points: {T}owards automatic argument
  summarization.
\newblock In Jurafsky, D., Chai, J., Schluter, N., and Tetreault, J. (eds.),
  \emph{Proceedings of the 58th Annual Meeting of the Association for
  Computational Linguistics}, pp.\  4029--4039, Online, July 2020. Association
  for Computational Linguistics.
\newblock \doi{10.18653/v1/2020.acl-main.371}.
\newblock URL \url{https://aclanthology.org/2020.acl-main.371}.

\bibitem[Chen et~al.(2023)Chen, Wu, Wu, and Yang]{chen2023learning}
Chen, S., Wu, J., Wu, Y., and Yang, Z.
\newblock Learning to incentivize information acquisition: Proper scoring rules
  meet principal-agent model.
\newblock In \emph{International Conference on Machine Learning}, pp.\
  5194--5218. PMLR, 2023.

\bibitem[Chen \& Yu(2021)Chen and Yu]{CY-21}
Chen, Y. and Yu, F.-Y.
\newblock Optimal scoring rule design.
\newblock \emph{arXiv preprint arXiv:2107.07420}, 2021.

\bibitem[Clark et~al.(2019)Clark, Lee, Chang, Kwiatkowski, Collins, and
  Toutanova]{clark2019boolq}
Clark, C., Lee, K., Chang, M.-W., Kwiatkowski, T., Collins, M., and Toutanova,
  K.
\newblock Boolq: Exploring the surprising difficulty of natural yes/no
  questions.
\newblock In \emph{Proceedings of the 2019 Conference of the North American
  Chapter of the Association for Computational Linguistics: Human Language
  Technologies, Volume 1 (Long and Short Papers)}, pp.\  2924--2936, 2019.

\bibitem[Conitzer \& Sandholm(2003{\natexlab{a}})Conitzer and
  Sandholm]{Conitzer03Automated}
Conitzer, V. and Sandholm, T.
\newblock Automated mechanism design: Complexity results stemming from the
  single-agent setting.
\newblock In \emph{Proceedings of the 5th International Conference on
  {{Electronic}} Commerce}, {{ICEC}} '03, pp.\  17--24, New York, NY, USA,
  September 2003{\natexlab{a}}. Association for Computing Machinery.
\newblock ISBN 978-1-58113-788-0.
\newblock \doi{10.1145/948005.948008}.

\bibitem[Conitzer \& Sandholm(2003{\natexlab{b}})Conitzer and
  Sandholm]{Conitzer03Automateda}
Conitzer, V. and Sandholm, T.
\newblock Automated mechanism design for a self-interested designer.
\newblock In \emph{Proceedings of the 4th {{ACM}} Conference on {{Electronic}}
  Commerce}, {{EC}} '03, pp.\  232--233, New York, NY, USA, June
  2003{\natexlab{b}}. Association for Computing Machinery.
\newblock ISBN 978-1-58113-679-1.
\newblock \doi{10.1145/779928.779974}.

\bibitem[Conitzer \& Sandholm(2004)Conitzer and
  Sandholm]{Conitzer04Selfinterested}
Conitzer, V. and Sandholm, T.
\newblock Self-interested automated mechanism design and implications for
  optimal combinatorial auctions.
\newblock In \emph{Proceedings of the 5th {{ACM}} Conference on {{Electronic}}
  Commerce}, {{EC}} '04, pp.\  132--141, New York, NY, USA, May 2004.
  Association for Computing Machinery.
\newblock ISBN 978-1-58113-771-2.
\newblock \doi{10.1145/988772.988793}.

\bibitem[Curry et~al.(2022)Curry, Sandholm, and
  Dickerson]{Curry22Differentiable}
Curry, M., Sandholm, T., and Dickerson, J.
\newblock Differentiable {{Economics}} for {{Randomized Affine Maximizer
  Auctions}}.
\newblock In \emph{{International Joint Conference on Artificial Intelligence
  (IJCAI)}}, 2022.

\bibitem[Curry et~al.(2024)Curry, Thoma, Chakrabarti, McAleer, Kroer, Sandholm,
  He, and Seuken]{Curry24Automated}
Curry, M., Thoma, V., Chakrabarti, D., McAleer, S., Kroer, C., Sandholm, T.,
  He, N., and Seuken, S.
\newblock Automated {{Design}} of {{Affine Maximizer Mechanisms}} in {{Dynamic
  Settings}}.
\newblock \emph{Proceedings of the AAAI Conference on Artificial Intelligence},
  38\penalty0 (9):\penalty0 9626--9635, March 2024.
\newblock ISSN 2374-3468.
\newblock \doi{10.1609/aaai.v38i9.28819}.

\bibitem[Danz et~al.(2022)Danz, Vesterlund, and Wilson]{danz2022belief}
Danz, D., Vesterlund, L., and Wilson, A.~J.
\newblock Belief elicitation and behavioral incentive compatibility.
\newblock \emph{American Economic Review}, 112\penalty0 (9):\penalty0
  2851--2883, 2022.

\bibitem[Duan et~al.(2023)Duan, Sun, Chen, and Deng]{Duan23Scalablea}
Duan, Z., Sun, H., Chen, Y., and Deng, X.
\newblock A {{Scalable Neural Network}} for {{DSIC Affine Maximizer Auction
  Design}}.
\newblock \emph{Advances in Neural Information Processing Systems},
  36:\penalty0 56169--56185, December 2023.

\bibitem[D{\"u}tting et~al.(2024)D{\"u}tting, Feng, Narasimhan, Parkes, and
  Ravindranath]{Dutting22Optimal}
D{\"u}tting, P., Feng, Z., Narasimhan, H., Parkes, D.~C., and Ravindranath,
  S.~S.
\newblock Optimal auctions through deep learning: Advances in differentiable
  economics.
\newblock \emph{Journal of the ACM}, 71\penalty0 (1):\penalty0 1--53, 2024.

\bibitem[Feldman et~al.(2014)Feldman, Gravin, and
  Lucier]{feldman2014combinatorial}
Feldman, M., Gravin, N., and Lucier, B.
\newblock Combinatorial auctions via posted prices.
\newblock In \emph{Proceedings of the twenty-sixth annual ACM-SIAM symposium on
  Discrete algorithms}, pp.\  123--135. SIAM, 2014.

\bibitem[Fu et~al.(2024)Fu, Ng, Jiang, and Liu]{fu2024gptscore}
Fu, J., Ng, S.~K., Jiang, Z., and Liu, P.
\newblock Gptscore: Evaluate as you desire.
\newblock In \emph{Proceedings of the 2024 Conference of the North American
  Chapter of the Association for Computational Linguistics: Human Language
  Technologies (Volume 1: Long Papers)}, pp.\  6556--6576, 2024.

\bibitem[Hackl et~al.(2023)Hackl, M{\"u}ller, Granitzer, and
  Sailer]{hackl2023gpt}
Hackl, V., M{\"u}ller, A.~E., Granitzer, M., and Sailer, M.
\newblock Is gpt-4 a reliable rater? evaluating consistency in gpt-4's text
  ratings.
\newblock In \emph{Frontiers in Education}, volume~8, pp.\  1272229. Frontiers
  Media SA, 2023.

\bibitem[Hartline et~al.(2023)Hartline, Shan, Li, and Wu]{HSLW-23}
Hartline, J.~D., Shan, L., Li, Y., and Wu, Y.
\newblock Optimal scoring rules for multi-dimensional effort.
\newblock In \emph{The Thirty Sixth Annual Conference on Learning Theory}, pp.\
   2624--2650. PMLR, 2023.

\bibitem[Hossain \& Okui(2013)Hossain and Okui]{hossain2013binarized}
Hossain, T. and Okui, R.
\newblock The binarized scoring rule.
\newblock \emph{Review of Economic Studies}, 80\penalty0 (3):\penalty0
  984--1001, 2013.

\bibitem[Hsu et~al.(2016)Hsu, Morgenstern, Rogers, Roth, and
  Vohra]{hsu2016prices}
Hsu, J., Morgenstern, J., Rogers, R., Roth, A., and Vohra, R.
\newblock Do prices coordinate markets?
\newblock In \emph{Proceedings of the forty-eighth annual ACM symposium on
  Theory of Computing}, pp.\  440--453, 2016.

\bibitem[Kimpara et~al.(2023)Kimpara, Frongillo, and
  Waggoner]{kimpara2023proper}
Kimpara, D., Frongillo, R., and Waggoner, B.
\newblock Proper losses for discrete generative models.
\newblock In \emph{International Conference on Machine Learning}, pp.\
  17015--17040. PMLR, 2023.

\bibitem[Kwiatkowski et~al.(2019)Kwiatkowski, Palomaki, Redfield, Collins,
  Parikh, Alberti, Epstein, Polosukhin, Kelcey, Devlin, Lee, Toutanova, Jones,
  Chang, Dai, Uszkoreit, Le, and Petrov]{gg-natural-questions}
Kwiatkowski, T., Palomaki, J., Redfield, O., Collins, M., Parikh, A., Alberti,
  C., Epstein, D., Polosukhin, I., Kelcey, M., Devlin, J., Lee, K., Toutanova,
  K.~N., Jones, L., Chang, M.-W., Dai, A., Uszkoreit, J., Le, Q., and Petrov,
  S.
\newblock Natural questions: a benchmark for question answering research.
\newblock \emph{Transactions of the Association of Computational Linguistics},
  2019.

\bibitem[Li et~al.(2022)Li, Hartline, Shan, and Wu]{LHSW-22}
Li, Y., Hartline, J.~D., Shan, L., and Wu, Y.
\newblock Optimization of scoring rules.
\newblock In \emph{Proceedings of the 23rd ACM Conference on Economics and
  Computation}, pp.\  988--989, 2022.

\bibitem[Lu et~al.(2024)Lu, Xu, Zhang, Kong, and Schoenebeck]{textual-pp}
Lu, Y., Xu, S., Zhang, Y., Kong, Y., and Schoenebeck, G.
\newblock Eliciting informative text evaluations with large language models.
\newblock \emph{the 25th ACM Conference on Economics and Computation}, 2024.

\bibitem[McCarthy(1956)]{mcc-56}
McCarthy, J.
\newblock Measures of the value of information.
\newblock \emph{Proceedings of the National Academy of Sciences of the United
  States of America}, 42\penalty0 (9):\penalty0 654, 1956.

\bibitem[M{\"o}bius et~al.(2022)M{\"o}bius, Niederle, Niehaus, and
  Rosenblat]{mobius2022managing}
M{\"o}bius, M.~M., Niederle, M., Niehaus, P., and Rosenblat, T.~S.
\newblock Managing self-confidence: Theory and experimental evidence.
\newblock \emph{Management Science}, 68\penalty0 (11):\penalty0 7793--7817,
  2022.

\bibitem[Morgenstern \& Roughgarden(2016)Morgenstern and
  Roughgarden]{Morgenstern16Learning}
Morgenstern, J. and Roughgarden, T.
\newblock Learning {{Simple Auctions}}.
\newblock In \emph{Conference on {{Learning Theory}}}, pp.\  1298--1318. PMLR,
  June 2016.

\bibitem[Morgenstern \& Roughgarden(2015)Morgenstern and
  Roughgarden]{Morgenstern2015Pseudo}
Morgenstern, J.~H. and Roughgarden, T.
\newblock On the pseudo-dimension of nearly optimal auctions.
\newblock In \emph{Advances in Neural Information Processing Systems}, 2015.

\bibitem[Neyman et~al.(2021)Neyman, Noarov, and Weinberg]{neyman2021binary}
Neyman, E., Noarov, G., and Weinberg, S.~M.
\newblock Binary scoring rules that incentivize precision.
\newblock In \emph{Proceedings of the 22nd ACM Conference on Economics and
  Computation}, pp.\  718--733, 2021.

\bibitem[Papireddygari \& Waggoner(2022)Papireddygari and Waggoner]{PW-22}
Papireddygari, M. and Waggoner, B.
\newblock Contracts with information acquisition, via scoring rules.
\newblock In \emph{Proceedings of the 23rd ACM Conference on Economics and
  Computation}, pp.\  703--704, 2022.

\bibitem[Rajpurkar et~al.(2016)Rajpurkar, Zhang, Lopyrev, and
  Liang]{rajpurkar-etal-2016-squad}
Rajpurkar, P., Zhang, J., Lopyrev, K., and Liang, P.
\newblock {SQ}u{AD}: 100,000+ questions for machine comprehension of text.
\newblock In Su, J., Duh, K., and Carreras, X. (eds.), \emph{Proceedings of the
  2016 Conference on Empirical Methods in Natural Language Processing}, pp.\
  2383--2392, Austin, Texas, November 2016. Association for Computational
  Linguistics.
\newblock \doi{10.18653/v1/D16-1264}.
\newblock URL \url{https://aclanthology.org/D16-1264}.

\bibitem[Roughgarden \& Schrijvers(2016)Roughgarden and
  Schrijvers]{Roughgarden16Ironing}
Roughgarden, T. and Schrijvers, O.
\newblock Ironing in the {{Dark}}.
\newblock In \emph{Proceedings of the 2016 {{ACM Conference}} on {{Economics}}
  and {{Computation}}}, {{EC}} '16, pp.\  1--18, New York, NY, USA, July 2016.
  Association for Computing Machinery.
\newblock ISBN 978-1-4503-3936-0.
\newblock \doi{10.1145/2940716.2940723}.

\bibitem[Sandholm et~al.(2007)Sandholm, Conitzer, and
  Boutilier]{Sandholm07Automated}
Sandholm, T.~W., Conitzer, V., and Boutilier, C.
\newblock Automated {{Design}} of {{Multistage Mechanisms}}.
\newblock In \emph{{International Joint Conference on Artificial Intelligence
  (IJCAI)}}, 2007.

\bibitem[Savage(1971)]{sav-71}
Savage, L.~J.
\newblock Elicitation of personal probabilities and expectations.
\newblock \emph{Journal of the American Statistical Association}, 66\penalty0
  (336):\penalty0 783--801, 1971.

\bibitem[Schneider et~al.(2023)Schneider, Schenk, Niklaus, and
  Vlachos]{SSNV-23}
Schneider, J., Schenk, B., Niklaus, C., and Vlachos, M.
\newblock Towards llm-based autograding for short textual answers.
\newblock \emph{arXiv preprint arXiv:2309.11508}, 2023.

\bibitem[Shen et~al.(2019)Shen, Tang, and Zuo]{Shen21Automateda}
Shen, W., Tang, P., and Zuo, S.
\newblock Automated {{Mechanism Design}} via {{Neural Networks}}.
\newblock In \emph{{Proceedings of the 18th International Conference on
  Autonomous Agents and MultiAgent Systems (AAMAS)}}, 2019.

\bibitem[Wang et~al.(2024)Wang, Jiang, and Parkes]{Wang24GemNet}
Wang, T., Jiang, Y., and Parkes, D.~C.
\newblock {{GemNet}}: {{Menu-Based}}, {{Strategy-Proof Multi-Bidder Auctions
  Through Deep Learning}}.
\newblock In \emph{Proceedings of the 2024 {{ACM Conference}} on {{Economics}}
  and {{Computation}}}, 2024.

\bibitem[Wu \& Hartline(2024)Wu and Hartline]{WH-24}
Wu, Y. and Hartline, J.
\newblock Elicitationgpt: Text elicitation mechanisms via language models.
\newblock \emph{working paper}, 2024.

\bibitem[Zheng et~al.(2023)Zheng, Chiang, Sheng, Zhuang, Wu, Zhuang, Lin, Li,
  Li, Xing, et~al.]{zheng2023judging}
Zheng, L., Chiang, W.-L., Sheng, Y., Zhuang, S., Wu, Z., Zhuang, Y., Lin, Z.,
  Li, Z., Li, D., Xing, E., et~al.
\newblock Judging llm-as-a-judge with mt-bench and chatbot arena.
\newblock \emph{Advances in Neural Information Processing Systems},
  36:\penalty0 46595--46623, 2023.

\end{thebibliography}
\bibliographystyle{icml2025}

\clearpage
\appendix
\section{Implementation Details}\label{app:implement}

In this section, we provide a detailed description of how we implement our methods and conduct the experiments, including the prompts and other parameters for LLM calls, the numerical solution to the convex optimization problem, as well as the pre/post-processing of human feedback.

\subsection{LLM Calls}
We use the \textsf{gemini-2.5} series models as the LLM oracles in our experiments. Specifically, we experiment with \textsf{gemini-2.5-flash-preview-04-17} for all tasks other than clustering the negative/potitive pairs. For clustering, we employed \textsf{gemini-2.5-pro-preview-05-06} due to its proficiency in handling long contexts. While calling LLMs, we set the temperature to $0$, the ``thinking'' feature disabled, and maximum output token $8192$. Next, we will provide a detailed description of each prompt used.

\subsubsection{Summarization Oracle}

The implementation of the summarization oracle includes three steps: summarizing instructor review, preparing negative/positive statement pairs from reviews, and clustering negative/positive statement pairs.

\begin{tcolorbox}[enhanced, colback=gray!7, frame hidden, parbox=false, before upper={\setlength{\parindent}{0pt}\setlength{\parskip}{0.46ex}\obeylines}]\begin{scriptsize}
\begin{small}\textbf{Summarizing Instructor Review}\end{small}

You are an AI assistant specializing in analyzing assignment reviews. Your task is to extract all evaluative points from a given review comment.

\textless{}review\_comment\textgreater{}{{REVIEW\_COMMENT}}\textless{}/review\_comment\textgreater{}

Please follow these steps to analyze the review comment:

1. Carefully read the entire review comment.

2. Extract all evaluative statements from the review. These should be comments that assess the quality, strengths, weaknesses, and suggestions. Ignore purely descriptive or meaningless statements. Ignore statements purely about specific scores and ratings.

3. Create an indexed list of these evaluative statements. Each entry should be a single sentence in a single line containing a distinct evaluation from the review.
- You should clearly convey the sentiment behind an evaluative statement.

4. After creating the indexed list. Split and Rewrite each evaluative statement into several abstract and concise statements, abandoning the specific expression.
- Make your entry abstract and concise.
- Always use "part A / B / C" in the output to refer parts, even if the input says "part a / b / c" or "part 1 / 2 / 3".
- If an evaluative statement contains comments on multiple distinct aspects, they need to be listed as multiple entries.
Example: "I like the overall idea, but authors need to revise the presentation and experiments" have 3 different aspects, "The overall idea is good", "The presentation need revision", and "The experiments need revision".
Example: "Part A is correct and part B is wrong" have 2 different aspects, "Part A is correct", and "Part B is wrong".
- Ignore the unimportant positive parts of negative statements and the unimportant negative parts of positive statements.
- Each new entry inherits the index of the original entry, even if there are duplicate indexes.

Your output should be structured as follows:

\textless{}numbered\_entries\textgreater{}[List your numbered entries here, one per line]\textless{}/numbered\_entries\textgreater{}

\textless{}rewrited\_entries\textgreater{}[Rewrite each entry into an abstract and concise statement]\textless{}/rewrited\_entries\textgreater{}

\end{scriptsize}\end{tcolorbox}

\begin{tcolorbox}[enhanced, colback=gray!7, frame hidden, parbox=false, before upper={\setlength{\parindent}{0pt}\setlength{\parskip}{0.46ex}\obeylines}]\begin{scriptsize}
\begin{small}\textbf{Preparing Negative/Positive Statement Pair}\end{small}
You are tasked with creating opposite evaluative statements for a given list of evaluative statements. For each statement provided, you need to create a new statement that has the same content but expresses the opposite emotion or sentiment. 

In addition, you also need to output whether the sentiment of the original statement is positive or negative.

Guidelines for creating opposite evaluative statements:
1. Maintain the same subject matter and key elements of the original statement.
2. Change the emotional tone or sentiment to its opposite (e.g., positive to negative, approval to disapproval).
3. Use similar language structure when possible, but modify words to reflect the opposite sentiment.
4. Ensure the new statement is coherent and makes sense in isolation.
5. Make the new statement as concise as possible.

Here is the list of evaluative statements:

\textless{}evaluative\_statements\textgreater{}
{{EVALUATIVE\_STATEMENTS}}
\textless{}/evaluative\_statements\textgreater{}

For each statement in the list, create an opposite version following the guidelines above. Present your results in the following format:

\textless{}result\_1\textgreater{}
\textless{}original\textgreater{}[Original evaluative statement]\textless{}/original\textgreater{}
\textless{}sentiment\textgreater{}[Sentiment of the original evaluative statement]\textless{}/sentiment\textgreater{}
\textless{}opposite\textgreater{}[Your created opposite evaluative statement]\textless{}/opposite\textgreater{}
\textless{}/result\_1\textgreater{}

\textless{}result\_2\textgreater{}
...
\textless{}/result\_2\textgreater{}

...

Ensure that each opposite statement accurately reflects a reversal of sentiment while maintaining the core content of the original statement.
\end{scriptsize}\end{tcolorbox}

\begin{tcolorbox}[enhanced, colback=gray!7, frame hidden, parbox=false, before upper={\setlength{\parindent}{0pt}\setlength{\parskip}{0.46ex}\obeylines}]\begin{scriptsize}
\begin{small}\textbf{Clustering Statement Pairs}\end{small}

You will be given a list of opinion pairs, where each pair consists of a positive opinion and its corresponding negative opinion. Your task is to analyze these pairs and cluster them based on similarity. Follow these steps:

1. First, read the list of opinion pairs provided:

\textless{}opinion\_pairs\textgreater{}{{OPINION\_PAIRS}}\textless{}/opinion\_pairs\textgreater{}

2. Next, cluster the unique pairs based on their similarity in topic or theme in \textless{}clustering\textgreater{} tag. Pairs in the same cluster should address roughly the same aspects of the subject matter. Follow these steps:
1) You need to first draft a set of cluster descriptions in the \textless{}draft\textgreater{} tag. Each cluster description must be specific:
- You should cluster opinion pairs discussing different parts in different clusters.
- The description should clearly indicate the target of evaluation, avoiding terms like "overall" or "assignment" and instead using "the proof," "part A," or "the answer."
- The description should clearly specify the evaluation criteria, avoiding terms like "quality" and instead using "correctness," "clarity," or "detail."
2) Then, based on these descriptions, analyze the following aspects in the \textless{}analysis\textgreater{} tag:
- Splitting and merging clusters: Merge clusters that are redundant. Split clusters that contain more than one parts or aspects.
- New clusters: Look for opinions that are not covered by any existing cluster. Create a new cluster when at least two opinions fit it, and ignore any lone opinion that cannot be grouped.
- Specificity check: Ensure each cluster description includes specific evaluation criteria, rather than vague terms.
- Limit the number of clusters: Ensure the total number of clusters is between 10 and 12.
3) After completing this analysis, redefine the cluster descriptions based on your findings and repeat the entire process.
4) Perform this iteration a total of four times, wrapping the results of each iteration inside \textless{}epoch\_i\textgreater{} tags, where i represents the iteration number.

You should follow this output format:

\textless{}clustering\textgreater{}
\textless{}epoch\_1\textgreater{}
\textless{}draft\textgreater{}[Your draft cluster descriptions]\textless{}/draft\textgreater{}
\textless{}analysis\textgreater{}[Your analysis here]\textless{}/analysis\textgreater{}
\textless{}/epoch\_1\textgreater{}
\textless{}epoch\_2\textgreater{}
...
\textless{}/epoch\_2\textgreater{}
...
\textless{}/clustering\textgreater{}

3. Complete your final cluster descriptions. For each cluster, generate an opinion pair as the cluster representative.  
- Ensure the opinion pair discusses exactly the core idea of the cluster description.  
- The opinion pair should be brief and omit details.  
- Do not use "need" or "need not" in your opinion pair. Instead, express what was done or what was failed to be done.  
- Ensure the positive opinion and the negative opinion present exact opposing views.  
- It is not necessary to summarize all content. Focus only on evaluating the most important aspect, and avoid using "and" to connect different aspects.  
- Avoid using extreme words such as "excellent" and "awful."

You should follow this output format:

\textless{}clusters\textgreater{}
\textless{}cluster\_1\textgreater{}
\textless{}description\textgreater{}[The description of the cluster]\textless{}/description\textgreater{}
\textless{}representative\textgreater{}[Positive opinion] [Negative opinion]\textless{}/representative\textgreater{}
\textless{}/cluster\_1\textgreater{}

\textless{}cluster\_2\textgreater{}
...
\textless{}/cluster\_2\textgreater{}

...

\textless{}/clusters\textgreater{}

\end{scriptsize}\end{tcolorbox}

\subsubsection{Question-Answering Oracle}

We directly query LLM to identify whether the review $\review$ is positive or negative for the summary point $\statetxt$.

\paragraph{Input} One review $\review$ and a summary point $\statetxt$.

\paragraph{Output} Positive ($1$), negative ($0$), or NA ($\bot$).

\begin{tcolorbox}[enhanced, colback=gray!7, frame hidden, parbox=false, before upper={\setlength{\parindent}{0pt}\setlength{\parskip}{0.46ex}\obeylines}]\begin{scriptsize}
\begin{small}\textbf{Question-Answering Oracle}\end{small}

You are an AI assistant specializing in analyzing assignment reviews. Your task is to infer which of the given positive/negative opinions is correct based on the provided review comment. You will be given two inputs:

\textless{}review\_comment\textgreater{}{{REVIEW\_COMMENT}}\textless{}/review\_comment\textgreater{}

\textless{}opinion\_pairs\textgreater{}{{OPINION\_PAIRS}}\textless{}/opinion\_pairs\textgreater{}

The review comment is the text of the review that you need to analyze. The opinion pairs consist of several lines, each containing a positive evaluation and its corresponding negative evaluation.

For each opinion pair, follow these steps to analyze and conclude in \textless{}result\textgreater{} tag:
1. Reprint the index of the opinion pair in \textless{}index\textgreater{} tag.
2. Copy the text of the opinion pair in \textless{}opinion\_pair\textgreater{} tag.
3. Carefully read and understand both the positive and negative opinions.
4. List all possibly relevant statements in the comment one by one in the \textless{}statements\textgreater{} tag. For each relevant statement, determine whether it supports the positive opinion, the negative opinion, or neither, and specify whether the support is explicit or partial.
- Focus on the original meaning of the statement and avoid speculation as much as possible. 
Example: The correctness of the assignment refers to the accuracy of the final answer and does not include the reasoning process.
Example: The correctness of the proof / claim does not affect the correctness of the answer.
Example: The wrong proof / answer / reasoning does not affect clarity.
5. Apply the following rules to determine the final conclusion in the \textless{}rubric\textgreater{} tag:
- If only one direction is supported, classify as that direction, even if it is only partially supported.
- If their are conflicts, classify as the direction with stronger support.
- If no statement is relevant to the opinion pair, classify as "Neither". Avoid selecting "Neither" whenever possible.
- At the end of the rubric, explicitly state you choose "Positive", "Negative", or "Neither".
6. Restate your choice of whether the review supports the positive, the negative, or neither in the \textless{}conclusion\textgreater{} tag.
- Only contain "Positive", "Negative", or "Neither" in the tag! Do not use words like "Correct", "Incorrect", "Clear", "Unclear".

Present your analysis and conclusion for each opinion pair in the following format:

\textless{}result\textgreater{}
\textless{}index\textgreater{}[The index of the input opinion pair here]\textless{}/index\textgreater{}
\textless{}opinion\_pair\textgreater{}[Copy the input opinion pair here]\textless{}/opinion\_pair\textgreater{}
\textless{}statements\textgreater{}
Statement: [Statement 1]
Analysis: [Analysis for Statement 1]
Statement: [Statement 2]
Analysis: [Analysis for Statement 2]
...
\textless{}/statement\textgreater{}
\textless{}rubric\textgreater{}[Apply the rubric here]\textless{}/rubric\textgreater{}
\textless{}conclusion\textgreater{}[Positive / Negative / Neither]\textless{}/conclusion\textgreater{}
\textless{}/result\textgreater{}
\textless{}result\textgreater{}...\textless{}/result\textgreater{}
...
\end{scriptsize}\end{tcolorbox}

\subsubsection{LLM Score}

% We leverage LLM-outputed score as one of the two reference scores for peer review. We query language model \textsf{gemini-2.5-flash-preview-04-17} to compare the peer review against the instructor review.

\begin{tcolorbox}[enhanced, colback=gray!7, frame hidden, parbox=false, before upper={\setlength{\parindent}{0pt}\setlength{\parskip}{0.46ex}\obeylines}]\begin{scriptsize}
\begin{small}\textbf{LLM Score}\end{small}

You are an AI assistant specializing in educational assessment. Your task is to evaluate a peer review of a course assignment by comparing it to an instructor's review of the same assignment. You will analyze the alignment between the two reviews and assign a score from 0 to 10.

First, you will be given the instructor's review first and then the peer review to be evaluated.

To evaluate the peer review, follow these steps:

1. Identify the points in the instructor's review in the \textless{}evaluation\_process\textgreater{} tag. Express the same aspect across different parts as separate points. For each point in the instructor's review:
1) Reprint the text of this point from the instructor's review.
2) Judge whether the content of this point is subjective or objective.
    - Objective content includes factual assessments, such as the correctness of the assignment or proofs. - Subjective content includes aspects like clarity or style.
3) Identify the importance of this point:
   - Give more weight to critical elements like the correctness of the assignment or proofs.
   - Consider subjective elements and minor discrepancies less impactful on the overall score.
4) Extract all relevant text of this point from the peer review.
5) Assess the following aspects:
    a. Content: Does the peer review cover the same main topics of this key point? b. Accuracy: Are the peer reviewer's observations and critiques accurate when compared to the instructor's key point? c. Depth: Does the peer review provide an appropriate level of detail and insight?
6) Judge the overall quality of the peer review on this point.

2. According to your evaluation, offer a comprehensive assessment of this peer review in the \textless{}assessment\textgreater{} tag, supported by justification.
    - highlighting the alignments or misalignments between the peer review and the instructor's review.
    - Taking into account both the importance of each key point and the degree of alignment.

3. After the assessment, first provide your reasoning, then assign a score from 0 to 10 based on the rubric, enclosed in the \textless{}scoring\textgreater{} tag.
- 0-1: Totally wrong or meaningless review: The review is irrelevant, incoherent, or shows a complete misunderstanding of the material.
- 2-3: Poor review: The review demonstrates significant factual inaccuracies or fails to address essential key points.
- 4-6: Somewhat valuable review: The review contains clear errors or omissions, but partially aligns with the instructor's review on some important points.
- 7-9: Good review: The review largely aligns with the instructor's review on key points, with only minor inaccuracies or omissions.
- 10: Exceptional review: The review is highly consistent with the instructor's on both content and reasoning, with minimal flaws.

4. Output your final score again in the \textless{}final\_score\textgreater{} tag, with only the number.

Present your final evaluation in the following format:

\textless{}evaluation\_process\textgreater{}Point 1: [Description]
- Instructor's review: [Reprint text of this point from the instructor's review]
- Objective/subjective: [Reasoning first to judge whether the content of this point is subjective or objective]
- Importance: [Reasoning first to identify the importance of this point]
- Peer review: [Extract all relevant text of this point from the peer review]
- Assessment: [Assess the content, accuracy, and depth in detail]
- Quality: [Judge the quality of the peer review in relation to this point]
Point 2: [Description] ...\textless{}/evaluation\_process\textgreater{}

\textless{}assessment\textgreater{}[Your comprehensive assessment of this peer review]\textless{}/assessment\textgreater{}\textless{}scoring\textgreater{}[Your reasoning and the score for the peer review based on the rubric]\textless{}/scoring\textgreater{}\textless{}final\_score\textgreater{}[Output the final score]\textless{}/final\_score\textgreater{}

Here is your input:

\textless{}instructor\_review\textgreater{}{{INSTRUCTOR\_REVIEW}}\textless{}/instructor\_review\textgreater{}

\textless{}peer\_review\textgreater{}{{PEER\_REVIEW}}\textless{}/peer\_review\textgreater{}

\end{scriptsize}\end{tcolorbox}

\section{Additional Results}\label{app:addi}

This section presents experimental results that are omitted from the main text.

\subsection{LLM‐Judge Scores Using GPT}

In our primary experiments, we obtain LLM‐judge scores by querying the \textsf{gemini‐2.5‐flash‐preview‐04‐17} model to assess each peer review against its corresponding instructor review, according to a predefined scoring rubric.

\begin{figure}[h]
\centering
\includegraphics[width=\linewidth]{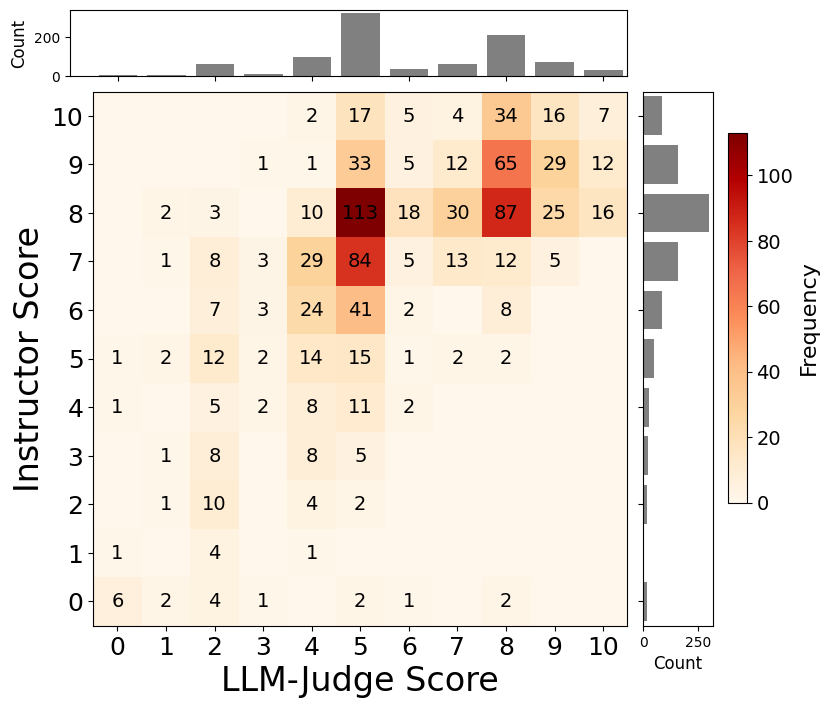}
\smallspac{3}
\caption{Joint distribution (instructor score vs. LLM-Judge score using \textsf{GPT-4.1})}
\label{fig:correlation_gpt}
\smallspac{3}
\end{figure}

To evaluate the robustness of this approach, we repeated the procedure using \textsf{GPT‐4.1} with the same prompt, thereby constructing a GPT‐based LLM‐judge. The resulting scores are shown in Figure~\ref{fig:correlation_gpt}. LLM-Judge with GPT shows a lower consistency with the instructor score. 

\begin{figure}[h]
\smallspac{2}
  \centering
  \subfigure[Instructor score vs.\ $\vsr$ aligned with instructor score.]{
    \includegraphics[scale = 0.3]{figs/joint_inst_asr.png}
    \label{fig:joint-inst}
  }
  \quad
  \subfigure[LLM-Judge score using \textsf{GPT-4.1} vs.\ $\vsr$ aligned with LLM-Judge score.]{
    \includegraphics[scale = 0.3]{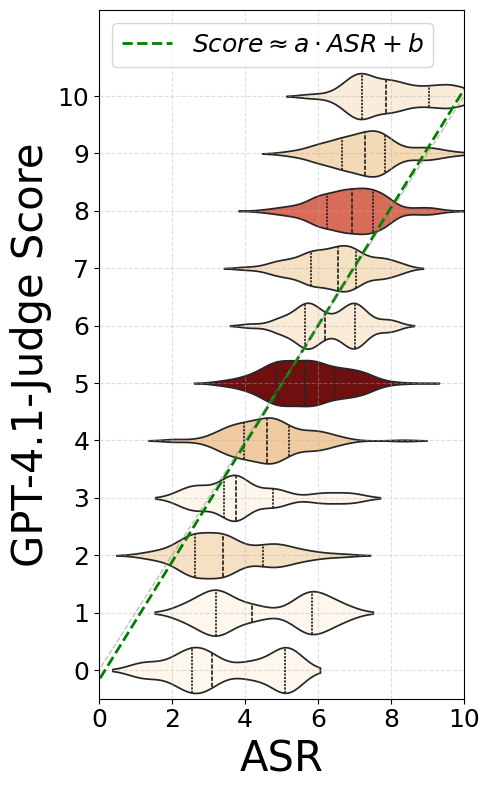}
  }
  \caption{Reference Scores vs. \vsr: The green dotted line represents the linear regression fitting reference score from $\vsr$. On both plots, the linear relationship is almost an identity function. }
  \label{fig:regression gpt}
  \smallspac{3}
\end{figure}

\Cref{fig:regression gpt} presents the same linear regression fitting the reference score from our $\vsr$. The regression line remains almost identical.

\end{document}